\documentclass[conference]{IEEEtran}
\IEEEoverridecommandlockouts
\usepackage{amsmath,amssymb,amsfonts}
\usepackage{algorithmic}
\usepackage{graphicx}
\usepackage{float}
\usepackage{textcomp}
\usepackage{color}
\usepackage[utf8]{inputenc}
\usepackage{multirow}
\usepackage[normalem]{ ulem }
\usepackage{soul}
\usepackage{dirtytalk}
\usepackage{atbegshi}
\AtBeginDocument{\AtBeginShipoutNext{\AtBeginShipoutDiscard}}

\newcommand{\R}{\mathbb{R}}

\newcommand{\Z}{\mathbb{Z}}

\newcommand{\q}[1]{``#1''}

\begin{document}

\title{A Topological \q{Reading} Lesson: Classification of MNIST using TDA}

\author{\IEEEauthorblockN{Ad\'elie Garin}
\IEEEauthorblockA{\textit{Laboratory for Topology and Neuroscience} \\
\textit{EPFL}\\
Lausanne, Switzerland \\
adelie.garin@epfl.ch}
\and
\IEEEauthorblockN{Guillaume Tauzin}
\IEEEauthorblockA{\textit{Laboratory for Topology and Neuroscience} \\
\textit{EPFL}
\\
Lausanne, Switzerland \\
guillaume.tauzin@epfl.ch}
}

\thanks{AG is supported by the Swiss National Science Foundation, grant number $CRSII5\_177237$ and GT by Innosuisse, grant number $32875.1$ $lP-ICT$.}

\maketitle

\begin{abstract}
We present a way to use Topological Data Analysis (TDA) for machine learning tasks on grayscale images. We apply persistent homology to generate a wide range of topological features using a point cloud obtained from an image, its natural grayscale filtration, and different filtrations defined on the binarized image. We show that this topological machine learning pipeline can be used as a highly relevant dimensionality reduction by applying it to the MNIST digits dataset. We conduct a feature selection and study their correlations while providing an intuitive interpretation of their importance, which is relevant in both machine learning and TDA. Finally, we show that we can classify digit images while reducing the size of the feature set by a factor 5 compared to the grayscale pixel value features and maintain similar accuracy. \end{abstract}


\section{Introduction}
Topological Data Analysis (TDA) \cite{topdata} applies techniques from algebraic topology to study and extract topological and geometric information on the shape of data. In this paper, we use persistent homology~\cite{perssurvey}, a tool from TDA that extracts features representing the numbers of connected components, cycles, and voids and their birth and death during an iterative process called a \emph{filtration}. Each of those features is summarized as a point in a \emph{persistence diagram}. Most commonly used on point clouds, persistent homology can also be used on images, using their voxel structure. One can define a filtration based on techniques from image processing, like dilation or erosion for example, and generate features that are intuitive, though otherwise difficult to extract. It has already been successfully applied to classification tasks such as in \cite{cubhom}, \cite{lida} and \cite{cubes} for example. 

 In this paper, we aim at harnessing the power of TDA for machine learning. We combine and compare a wide range of TDA techniques to extract features from images that are usually used separately. The more classical tools based on point clouds are integrated as well by treating each voxel as a point. We apply our TDA pipeline to the MNIST dataset to generate a set of topological features from each image. Using a feature selection algorithm, we highlight and provide an intuitive interpretation of the most important ones. Their systematic comparative study enables one to investigate the underlying characteristic shapes of a dataset of images. Besides, we also show that a pipeline of selected TDA techniques act as a very effective dimensionality reduction algorithm. Indeed, a set of $28$ topological features are sufficient to attain the same classification accuracy as the set of grayscale pixel values. 

We start by defining persistent homology on point clouds in Section~\ref{ptcloud}. We then describe how it can be used on images and introduce several filtrations of binary images in Section~\ref{pers_images}. In Section~\ref{classification} we explain how to extract machine learning-ready features from the persistence diagrams obtained and describe a generic TDA machine learning pipeline. In Section~\ref{features_imp} we present and interpret the features that were important for the classification of MNIST images and discuss the results in Section~\ref{supervised}.

\section{Persistent homology on point clouds}\label{ptcloud}
Persistent homology extracts the birth and death of topological features throughout a filtration built from a dataset. We start here by providing the necessary mathematical background to define a filtration from a point cloud and the extraction of its topological invariants. This first requires the notion of a simplicial complex.

\subsection{Simplicial complexes}
\begin{figure}[!t] 
    \centering
    \includegraphics[height=0.9cm]{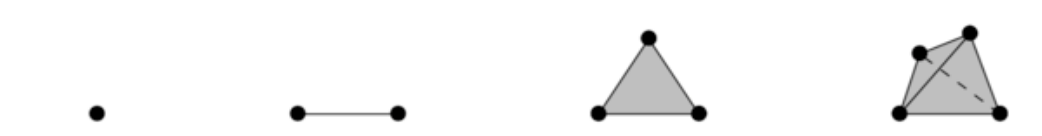}
    
    \hspace*{-0.2cm} \footnotesize $0$-simplex \hspace*{0.4cm} $1$-simplex \hspace*{0.8cm} $2$-simplex \hspace*{0.6cm} $3$-simplex
    \caption{Simplices of dimension $0$, $1$, $2$ and $3$.}
    \label{simp}
\end{figure}

Persistent homology on point clouds relies on objects referred as \emph{simplices}, which are the building blocks of the higher-dimensional counterparts of graphs called \emph{simplicial complexes}. A $k$-simplex $\sigma$ is the convex hull of $k+1$ affinely independent points (see Figure \ref{simp}). A $0$-simplex is also called a \textit{vertex} and a $1$-simplex an \textit{edge}.

A \emph{finite simplicial complex} is a finite collection of simplices satisfying some conditions. Its \textit{dimension} is the maximal dimension of its simplices.

From a dataset, one can obtain a simplicial complex using the Vietoris-Rips construction.
For $X$ a point cloud in $\R^n$, its \textit{Vietoris-Rips complex} \cite{Vietoris} of parameter $\varepsilon$, denoted by $VR(X,\varepsilon)$, is the simplicial complex with vertex set $X$ and where $\lbrace x_0,x_1,...,x_k \rbrace$ spans a $k$-simplex if and only if $d(x_i,x_j)\leq \varepsilon$ for all $0 \leq i,j \leq k$.

\subsection{Filtrations of simplicial complexes}
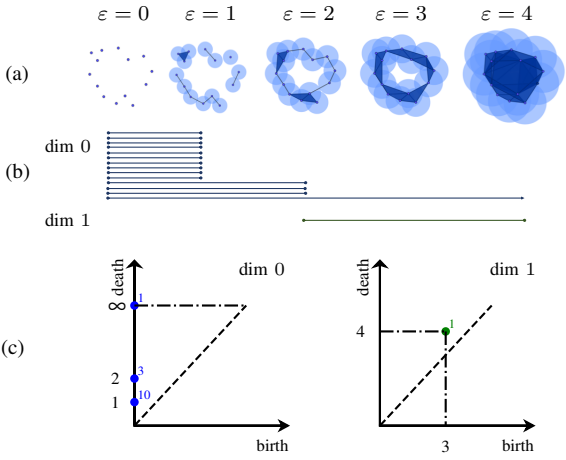
\begin{figure}[!t]
 
\centering
    
 \footnotesize
 
 \includegraphics[scale=0.2]{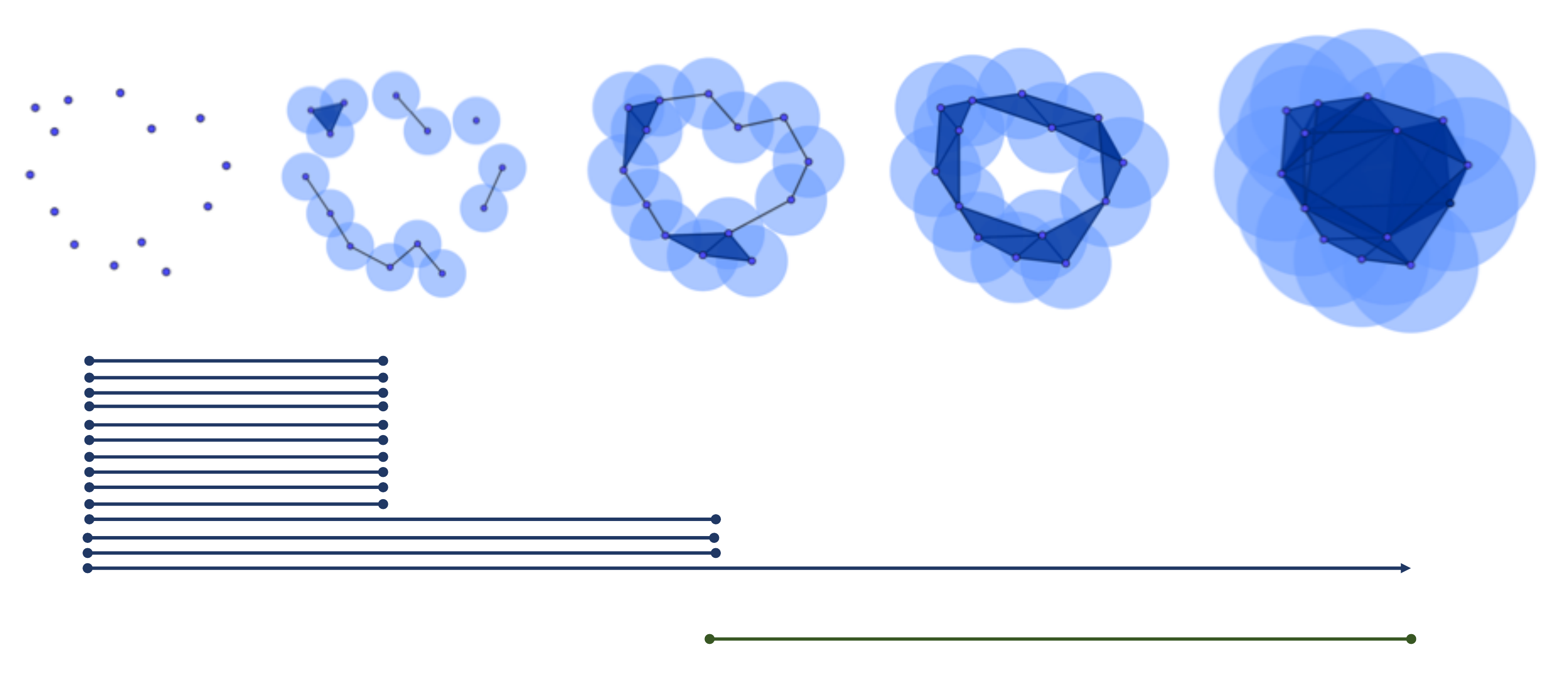}
 \put(-215,57){(a)}
  \put(-215,20){(b)}
  \put(-180,80){$\varepsilon = 0$}
  \put(-147,80){$\varepsilon = 1$}
  \put(-110,80){$\varepsilon = 2$}
  \put(-75,80){$\varepsilon = 3$}
  \put(-35,80){$\varepsilon = 4$}
  \put(-200,30){\scriptsize dim $0$}
  \put(-200,2){\scriptsize dim $1$}

 \includegraphics[scale=0.8]{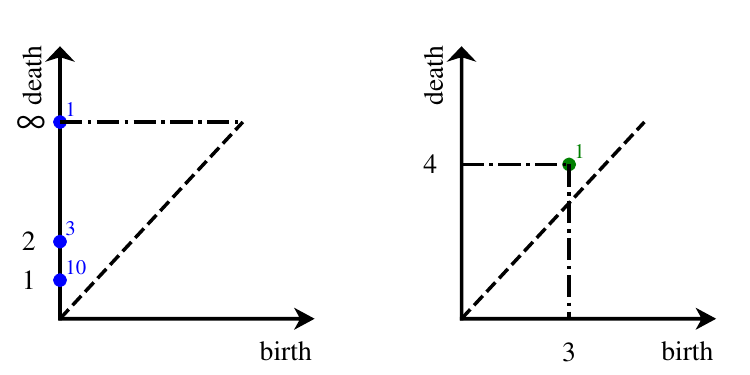}
   \put(-210,40){(c)}
     \put(-120,70){\scriptsize dim $0$}
  \put(-25,70){\scriptsize dim $1$}
    
    \caption{An example of a filtration (a), its barcode (b) and its persistence diagram (c). The $5$-step Vietoris-Rips filtration from a point cloud is shown in (a). We only take five values $\varepsilon=0,1,2,3$ and $\varepsilon=4$, where everything is filled in. 
    The birth-death pairs of connected components are indicated in blue and the $1$-cycle pair in green. 
    It is common to add the value $\infty$ to the diagram to indicate the cycles that never die. The index above each point corresponds to its \emph{multiplicity}: the number of times it appears in the barcode. For example, the point with coordinates $(0,1)$ corresponds to the $10$ bars of connected components that were born at time $0$ and died at time $\varepsilon=1$. The values indicated on the axis correspond to the filtration value $\varepsilon$.}
    \label{VR_filt}
\end{figure}

As $\varepsilon$ grows, so does the Vietoris-Rips complex of a point cloud. This defines a \emph{filtration} of simplicial complexes \textit{i.e.} a nested sequence of simplicial complexes  $\{VR(X;\varepsilon)\}_{\varepsilon \geq 0}$ satisfying $VR(X;\varepsilon_1) \subseteq VR(X;\varepsilon_2)$ if $\varepsilon_1 \leq \varepsilon_2$. 
 An example of a $5$-step filtration of Vietoris-Rips complexes is shown in Figure \ref{VR_filt}a. Balls are drawn around each point to represent the distance between them: if two balls of radius $\varepsilon$ intersect, the two points are at distance at most $2\varepsilon$.
 
\subsection{Persistent homology} \label{pers}

\section{Persistent homology on images} \label{pers_images}

The Vietoris-Rips filtration can also be applied directly to images, by seeing the pixels as a point cloud in a two-dimensional Euclidean space. However, a point cloud is not the most natural representation of an image. Images are made of pixels, or voxels in higher dimension, and thus have a natural grid structure that we can exploit. Hence, we use \emph{cubical complexes} in place of simplicial complexes, extending~\cite{cubhom} to the context of persistent homology.

\subsection{Images as cubical complexes}

\begin{figure}[!t]
    \centering
    \includegraphics[height=1cm]{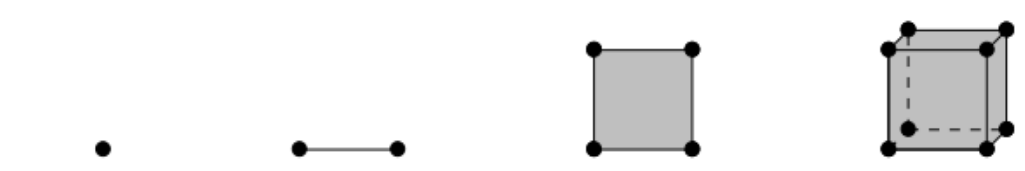}

     \footnotesize $0$-cube \hspace*{0.6cm} $1$-cube \hspace*{0.7cm} $2$-cube \hspace*{0.8cm} $3$-cube 
    \caption{Cubes of dimension $0$, $1$, $2$ and $3$.}
    \label{cubes}
\end{figure}

The cubical analog of a simplicial complex is a cubical complex, in which the role of simplices is played by cubes of different dimensions, as in Figure \ref{cubes}.
A finite \emph{cubical complex} in $\R^d$ is a union of cubes aligned on the grid $\Z^d$ satisfying some conditions similar to the simplicial complex case. 

Persistent homology (Section~\ref{pers}) is defined for filtrations of cubical complexes as well. Hence, we now introduce a way to represent images as cubical complexes and then explain how we build filtrations of cubical complexes from binary images. 

A \emph{$d$-dimensional image} is a map $\mathcal{I} : I \subseteq \Z^d \longrightarrow \R.$ An element $v \in I$ is called a \emph{voxel} (or \emph{pixel} when $d=2$) and the value  $\mathcal{I}(v)$ is called its \emph{intensity} or \emph{greyscale value}. In the case of \emph{binary images}, which are made of only black and white voxels, we consider a map $\mathcal{B} : I \subseteq \Z^d \longrightarrow \{0,1\}$. In a slight abuse of terminology, we call the subset $I \subseteq \Z^d$ an image.

There are several ways to represent images as cubical complexes. In~\cite{cubes}, the voxels are represented by vertices and cubes are built between those vertices. Here, we choose the approach in which a voxel is represented by a $d$-cube and all of its faces (adjacent lower-dimensional cubes) are added. We get a function on the resulting cubical complex $K$ by extending the values of the voxels to all the cubes $\sigma$ in $K$ in the following way: \[\mathcal{I}' (\sigma) :=  \min_{\sigma \text{ face of } \tau}\mathcal{I}(\tau). \]

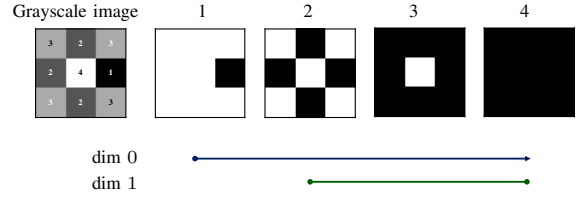
\begin{figure}[!t]
    \centering
    \scriptsize
    \includegraphics[scale=0.35]{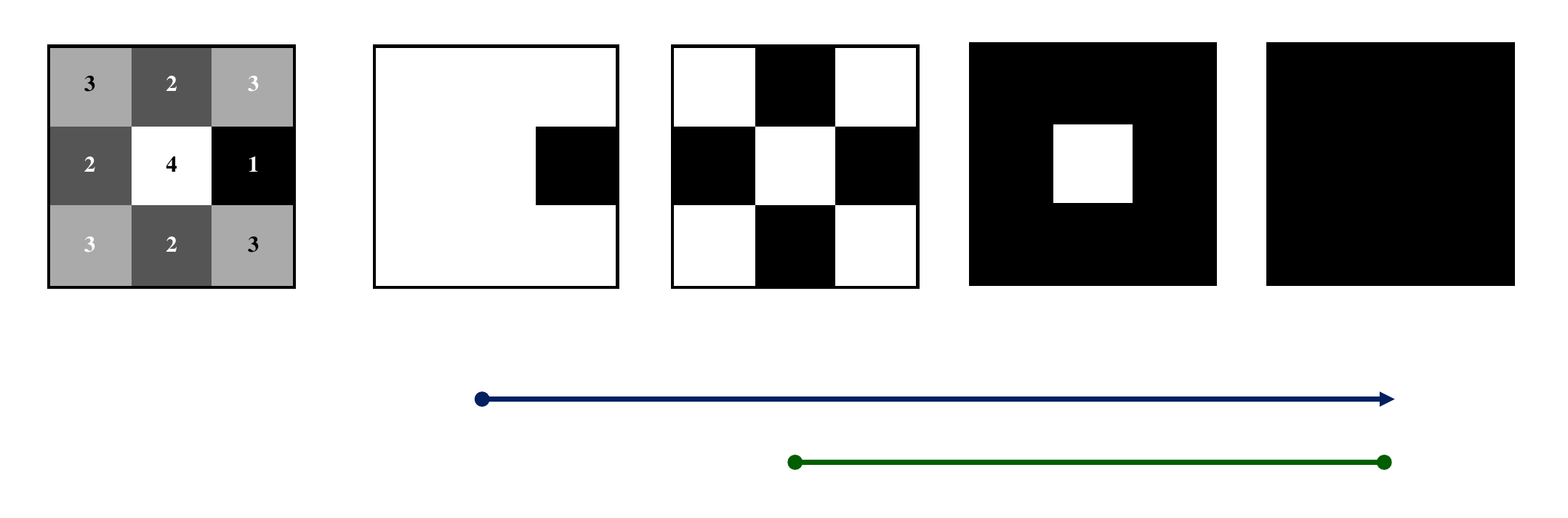}
    \put(-220,70){Grayscale image}
    \put(-150,70){1}
    \put(-110,70){2}
    \put(-70,70){3}
    \put(-28,70){4}
    \put(-190,15){dim $0$}
    \put(-190,5){dim $1$}

    \caption{A filtration of cubical complexes induced by the grayscale image on the left and its corresponding barcode. We see only one connected component that appears at time $1$ and survives until the end, as all the new pixels are always connected to the previous ones. A $2$-cycle appears at time $4$ and gets filled in at time $4$.}
    \label{barcode}
\end{figure}

A grayscale image comes with a natural filtration embedded in the grayscale values of its pixels. Let $K$ be the cubical complex built from the image $I$. Let $$K_i := \{\sigma \in K \mid \mathcal{I}'(\sigma) \leq i\}$$ be the $i$-th \emph{sublevel set} of $K$. The set $\{K_i\}_{i \in \text{Im}(I)}$ defines a filtration of cubical complexes, indexed by the value of the grayscale function $\mathcal{I}$.  The steps we factor through to obtain a filtration from a grayscale image are then:
$$\text{Image} \rightarrow \text{Cubical complex} \rightarrow \text{Sublevel sets} \rightarrow \text{Filtration}. $$
Figure \ref{barcode} shows an example of a grayscale image filtration.

\subsection{Filtrations of binary images}

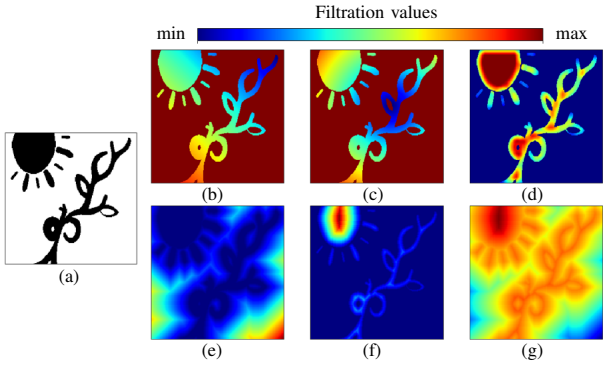
\begin{figure}[!t]
    \centering

\vspace{0.1cm}
  
\includegraphics[scale=0.25]{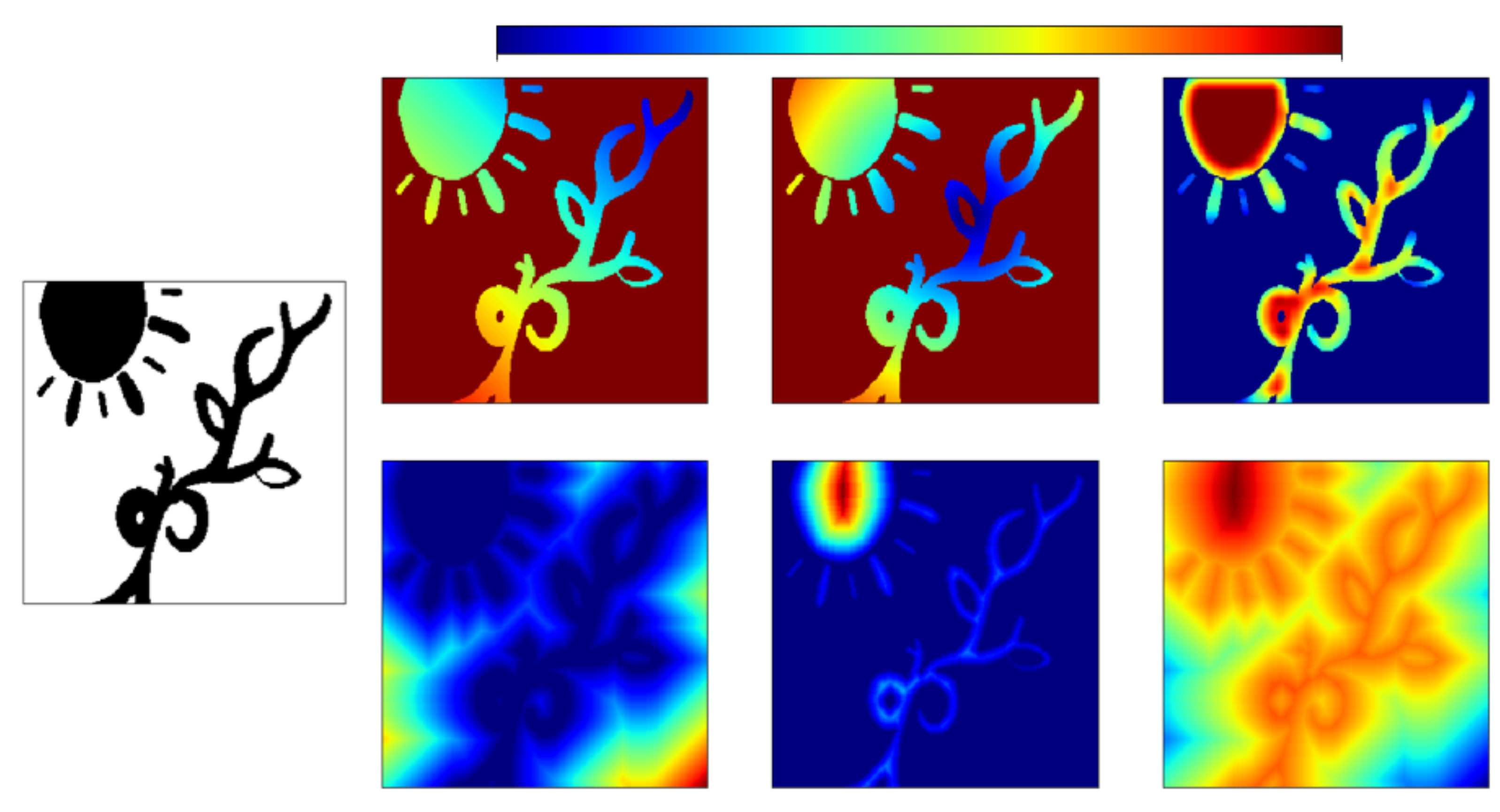}
\scriptsize 
\put(-207,25){(a)}
\put(-153,56){(b)}
\put(-92,56){(c)}
\put(-32,56){(d)}
\put(-153,-3){(e)}
\put(-92,-3){(f)}
\put(-32,-3){(g)}
\put(-170,117){min}
\put(-19,117){max}
\put(-110,125) {Filtration values}
\vspace{0.2cm}    
    \caption{Grayscale images obtained from the different types of filtrations. For visibility reasons, we choose a colored map to represent grayscale values. The original binary image we use to illustrate our filtrations is shown in (a) and also gives a binary filtration. The height filtration with vector $(-1,-1)$ is shown in (b), the radial filtration with center in the blue leaf in (c), the density filtration in (d), the dilation filtration in (e), the erosion filtration in (f) and the signed distance filtration in (g).}
    \label{images}
\end{figure}

Let $\mathcal{B}: I \subseteq \Z^d \longrightarrow \{0,1\}$ be a binary image. By building grayscale filtrations from it, we can highlight topological features. In this section, we propose various techniques to transform it into a grayscale image which we filter as defined in the previous section. We illustrate our filtrations on a specific example shown in Figure \ref{images}. 

\subsubsection{Binary filtration}\label{binary}
The binary filtration consists of computing the persistence diagram straight from the binary image, \emph{i.e.} by considering the binary values of the pixels as a two-level filtration. It is related to computing the homology of the image.

\subsubsection{Height filtration}
The height filtration is inspired by Morse theory and by the Persistent Homology Transform \cite{PHT}. For cubical complexes, we define the \emph{height filtration} $\mathcal{H}: I \longrightarrow \R$ of a $d$-dimensional binary image $I$ by choosing a direction $v \in \R^d$ of norm $1$ and defining new values on all the voxels of value $1$ as follows: if $p \in I$ is such that $\mathcal{B}(p) =1$, then one assigns a new value $\mathcal{H}(p) := <p,v>$, the distance of $p$ to the hyperplane defined by $v$. If $\mathcal{B}(p) = 0$, then $\mathcal{H}(p):= H_{\infty}$, where $H_{\infty}$ is the filtration value of the pixel that is the farthest away from the hyperplane.

\subsubsection{Radial filtration}
The \emph{radial filtration} $\mathcal{R}$ of $I$ with center $c \in I$, inspired from \cite{lida}, is defined by assigning to a voxel $p$ the value corresponding to its distance to the center
\[ \mathcal{R}(p):=
\begin{cases}
      \Vert c - p \Vert_2  \quad \text{ if }\mathcal{B}(p)=1\\
       \mathcal{R}_\infty \quad \quad \text{ if  } \mathcal{B}(p)=0.\\
       \end{cases} \]
where $\mathcal{R}_\infty$ is the distance of the pixel that is the farthest away from the center.

\subsubsection{Density filtration}
The density filtration gives each voxel a value depending on the number of neighbors it has at a certain distance. For a parameter $r$, the radius of the ball we want to consider, the density filtration is: $$\mathcal{D}_e(p) := \#\{v \in I, \mathcal{B}(v)=1 \text{ and } \Vert p-v \Vert \leq r \},$$ where the norm can be any norm on $\R^d$, but we choose the $L1$-norm in our implementation.

\subsubsection{Dilation filtration}
The \emph{dilation filtration} of $I$ defines a new grayscale image $\mathcal{D}: I \longrightarrow \R$ as follows: a vertex $p$ in $I$ is assigned the smallest distance to a vertex of value $1$ in $I$: $$\mathcal{D}(p):= \min \{ \Vert p - v \Vert_1, \text{  }  \mathcal{B}(v) = 1 \}.$$

\subsubsection{Erosion filtration}
The erosion filtration is the inverse of the dilation filtration. Instead of dilating the object it erodes it. To obtain the erosion filtration, one applies the dilation filtration to the inverse image, where $0$ and $1$ are switched. Note that in the case of extended persistence \cite{extended}, the two filtrations return the same information by duality, but it is not the case here as we consider only standard persistence.

\subsubsection{Signed distance filtration}
The signed distance filtration is a combination of the erosion and dilation filtrations that returns both positive and negative values. It takes positive values on the $1$-valued voxels by taking the distance to the boundary and negative values on the $0$-valued voxels, by attributing them the negative distance to the same boundary.

\section{Supervised learning on the MNIST dataset}

The MNIST dataset consists of $50,000$ images of handwritten digits of size $28 \times 28$. In this section, we study this dataset using a machine learning TDA pipeline based on the techniques introduced above. We compare the importance of our topological features and use them to classify the MNIST digits.

\subsection{Topological pipeline for Machine Learning}\label{classification}

We propose a generic machine learning pipeline to generate persistent homology features from images. Starting from a grayscale image, we use the voxel values directly as a filtration. To be able to make use of the binary filtrations described previously, we binarize the image according to a pixel value threshold of $40\%$. For the height and radial filtrations, we use each of the directions and centers shown in Figure \ref{features}a in order to cover uniformly the set of possible directions and centers. As for the density filtrations, we use $r=2, \, 4, \, 6$.

Furthermore, we can view each pixel as a point in a two-dimensional Euclidean space and consider the associated Vietoris-Rips filtration. From each of these filtrations, we compute the persistence diagrams that describe connected components (dimension 0) and 1-cycles (dimension 1). A note for topologists: we use coefficients in $\Z_{2}$.

\begin{figure}[!t]
    \centering
    \includegraphics[scale=0.1]{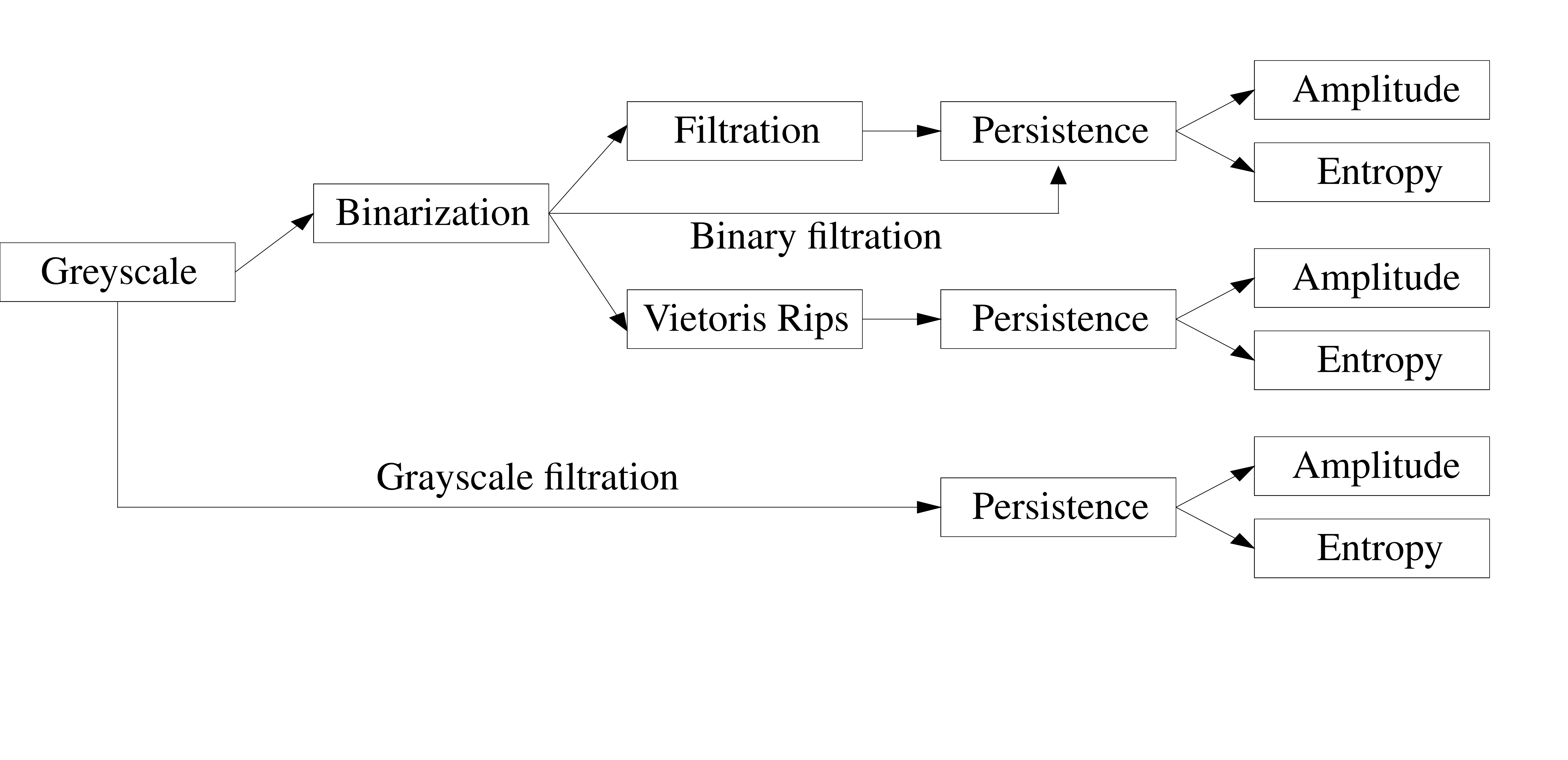}
    \vspace*{-0.8cm}
    \caption{The pipeline to classify grayscale images. The resulting feature vector has $728$ components.}
    \label{pipeline}
\end{figure}

 Each of these operations generates a persistence diagram. The strength of persistent homology comes from the \emph{stability theorem}~\cite{stab}: if two objects are similar, then the resulting persistence diagrams will be close as well. We can, therefore, measure how close two images are topologically speaking by comparing signature real numbers extracted from their persistence diagrams. There are many ways to do so; see~\cite{TDAML} for a survey of the different methods. In this work, we propose to use the \emph{amplitude} of each persistence diagram or its \emph{entropy}. The different steps for our topological pipeline taking as inputs images and generating persistence features are shown in Figure \ref{pipeline}. 

We define the \emph{amplitude} of a persistence diagram as its distance to the empty diagram, which contains only the diagonal points. In this paper, we use two metrics (the Wasserstein and Bottleneck distances) and three kernels (Betti curves, persistence landscapes~\cite{landscape} and heat kernel~\cite{heat}), for which diagrams are sampled using $100$ filtration values and the amplitude of the kernel is obtained using either $L1$ or $L2$ norms. 

\subsubsection{Betti curves} \label{betti}

The \emph{Betti curve} $B_n : I \longrightarrow \R$ of a barcode $D= \{ (b_j,d_j) \}_{j \in I}$ is the function that return for each step $i \in I$, the number of bars $(b_j,d_j)$ that contains $i$:
$$ i \mapsto \# \{(b_j,d_j), i \in (b_j,d_j) \}.$$ 

\subsubsection{Persistence landscapes}
Introduced in~\cite{landscape}, the $k$-th persistence landscape of a barcode $\{(b_i,d_i)\}_{i=1}^n$ is the function $\lambda_k : \R \longrightarrow [0,\infty)$ where $\lambda_k(x)$ is the $k$-th largest value of $\{f_{(b_i,d_i)}(x) \}_{i=1}^n$, with

\[f_{(b,d)}(x) =
\begin{cases}
      0  \quad \text{ if } x \notin (b,d)\\
       x-b \quad \text{ if } x \in (b,\frac{b+d}{2})\\
       -x+d \quad \text{ if } x \in (\frac{b+d}{2},d).
       \end{cases} \]
The parameter $k$ is called the layer. Here we consider curves obtained by setting $k= 1$ and $k\in\{1,2\}$. 

\subsubsection{Heat kernel}
In~\cite{heat}, the authors introduce a kernel by placing Gaussians of standard deviation $\sigma$ over every point of the persistence diagram and a negative Gaussian of the same standard deviation in the mirror image of the points across the diagonal. The output of this operation is a real-valued function on $\R^2$.  In this work, we consider two possible values for $\sigma$, $10$ and $15$ (in the unit of discrete filtration values).

\subsubsection{Wasserstein amplitude}
Based on the \emph{Wasserstein distance}, the \emph{Wasserstein amplitude} of order $p$ is the $Lp$ norm of the vector of point distances to the diagonal:

$$A_W= \frac{\sqrt{2}}{2}(\sum_{i} (d_i- bi)^p  )^{\frac{1}{p}}. $$
In this paper, we use $p=1,2$. 

\subsubsection{Bottleneck distance}
When we let $p$ go to $\infty$ in the definition of the Wasserstein amplitude, we obtain the \emph{Bottleneck amplitude}:
$$ A_B= \frac{\sqrt{2}}{2}\sup_{i}  (d_i- b_i). $$

\subsubsection{Persistent entropy}
The \emph{persistent entropy} of a barcode, introduced in~\cite{entropy}, is a real number extracted by taking the Shannon entropy of the persistence (lifetime) of all cycles: 
\[ PE(D) = \sum_{i=1}^n \frac{l_i}{L(B)} \log (\frac{l_i}{L(B)}),\] where $l_i := d_i-b_i$ and $L(B) := l_1 + ,... + l_n$ is the sum of all the persistences.

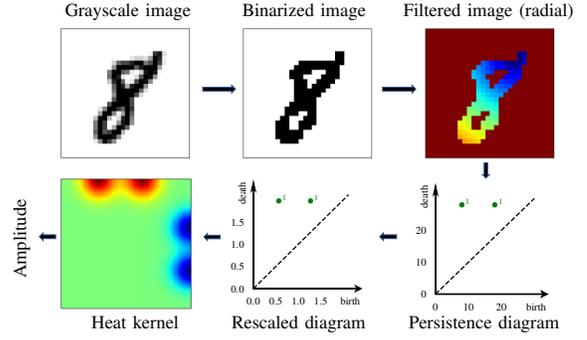
\begin{figure}[!t]
    \centering
    \scriptsize
    \includegraphics[scale=0.22]{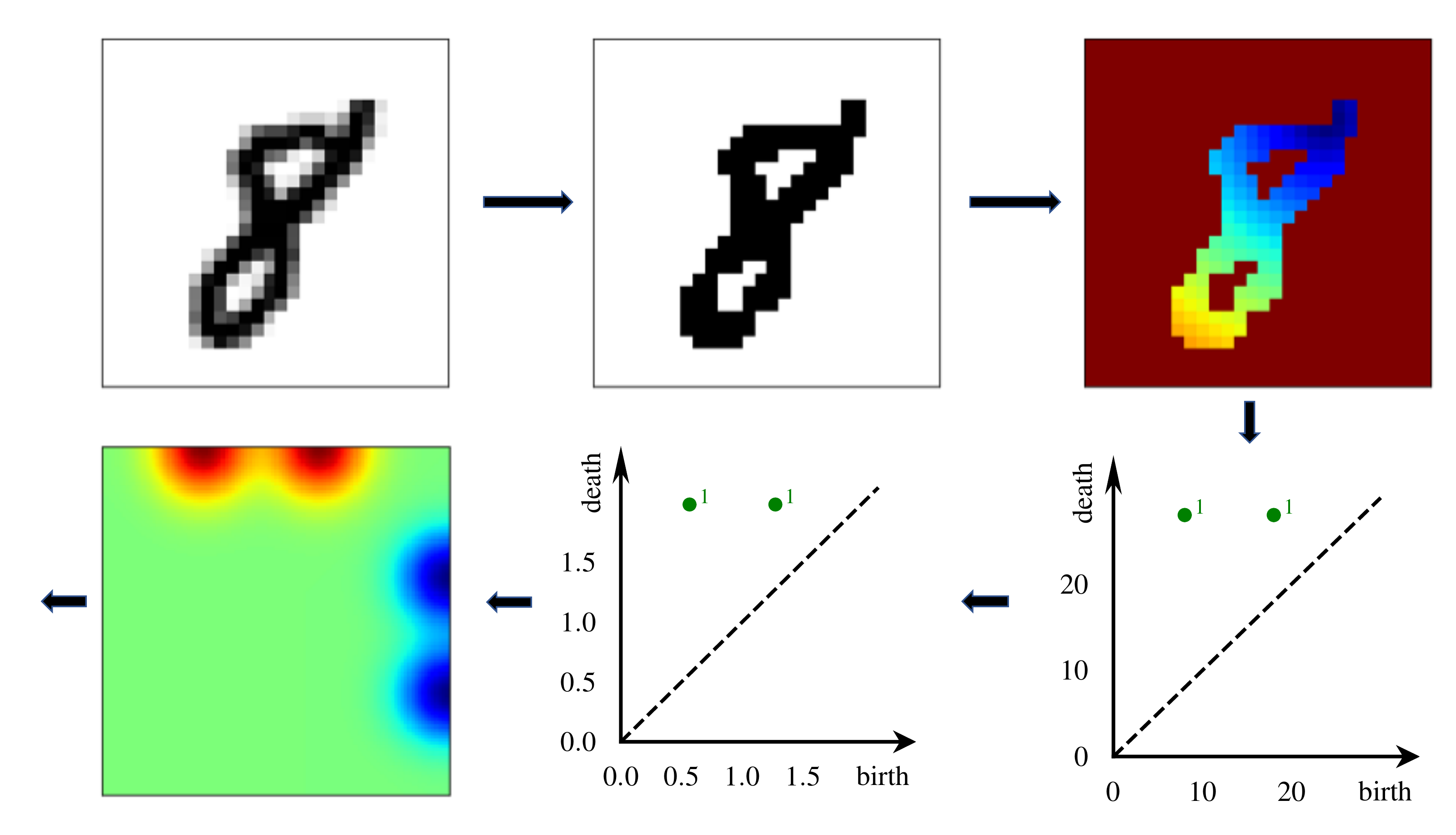}
    \put(-210,17){ \rotatebox{90}{Amplitude}}
    \put(-188,116){Grayscale image}
    \put(-121,116){Binarized image}
    \put(-60,116){Filtered image (radial)}
     \put(-178,-2){Heat kernel}
    \put(-125,-2){Rescaled diagram}
    \put(-58,-2){Persistence diagram}
    \caption{An example of our pipeline on a MNIST image with a choice of filtration (radial, center = (20, 6)), for the subdiagram of dimension 1 and a choice of feature generator (heat kernel, $\sigma=10$).}
    
    \label{ex_pipeline}
\end{figure}

Note that before we calculate the amplitude of any diagrams, we rescale it by the maximal Bottleneck amplitude of the diagrams obtained by the same filtration of the whole collection of images. We show one of the topological pipelines applied to an MNIST digit in Figure \ref{ex_pipeline}. 

\subsection{Features importance} \label{features_imp}
\begin{figure}[!t]
    \centering
    \includegraphics[scale=0.35]{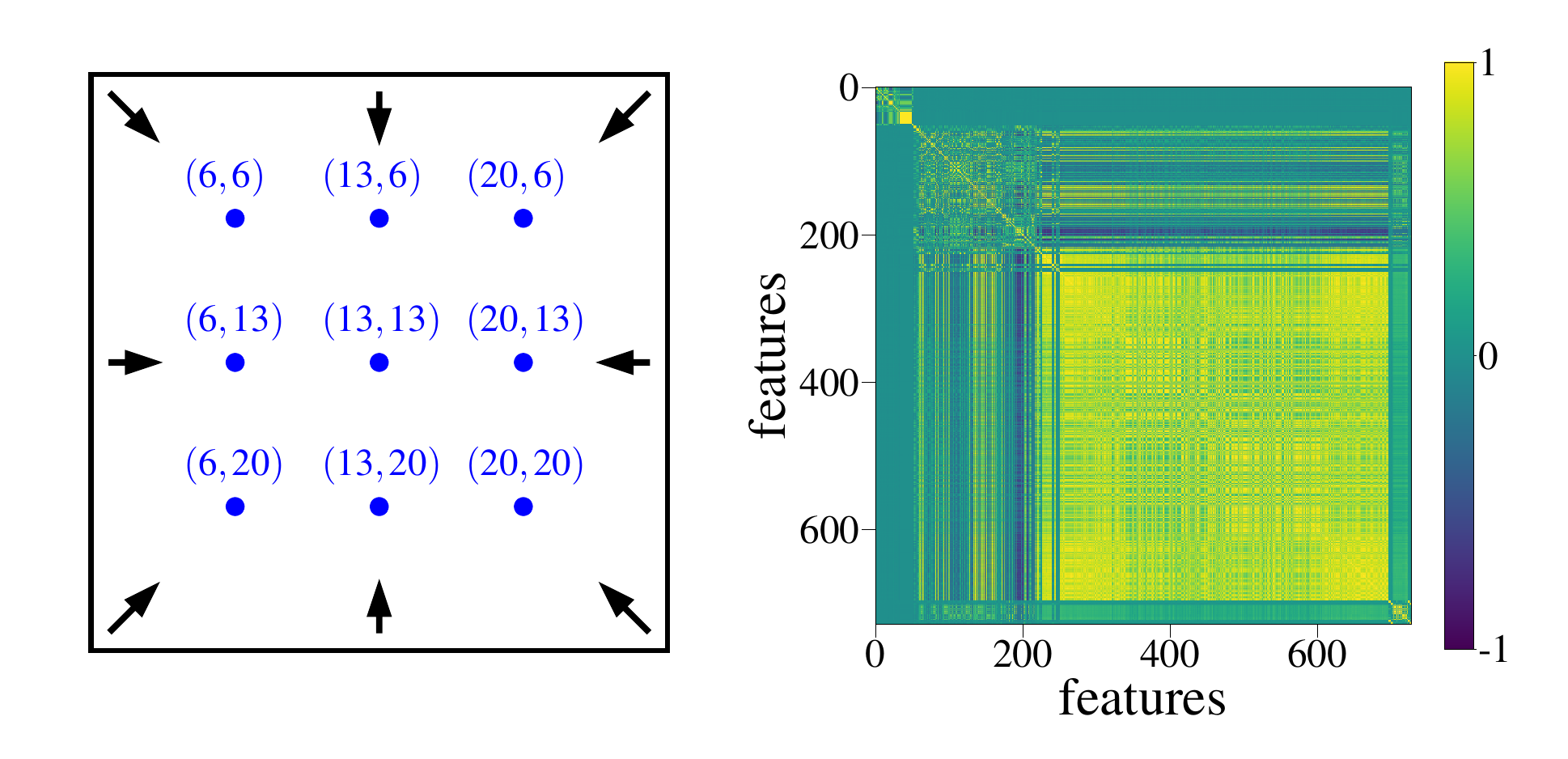}
    \vspace{0.2cm}
    \put(-155,-7){\scriptsize (a)}
    \put(-55,-7){\scriptsize (b)}
    \caption{The directions and centers that we used for the height and radial filtrations of the pipeline (a) and the Pearson correlation matrix between the features ordered by importance (b).}
    \label{features}
\end{figure}

We first look at feature importance using the random forest algorithm with $10,000$ trees on a training set of $40,000$ images. Our pipeline generates $728$ features, which can be correlated~\cite{Gard}. This is encoded in the Pearson correlation matrix in Figure~\ref{features}b where a large number of correlation values are very close to $1$.
Therefore, we select only the most important features that are not fully correlated, \emph{i.e.} based on a Pearson correlation strictly smaller than $0.9$. The resulting list of new features contains $84$ features instead of $728$. 

We describe the top eight of the $84$ selected features in Table~\ref{bestof_decor}.  
The height filtrations appear to lead to significant topological signatures. Along with the radial filtration, they provide global information about the number of connected components as well as their positions in the image. For example, a $6$ and a $9$ can be easily differentiated with height filtrations of directions $[1,0]$ and $[-1, 0]$. Overall, the study of features' importance from this generic TDA pipeline provide a systematic approach to understanding the shapes on the images of the dataset.
Persistent entropy dominates diagram vectorization methods. It was designed to extract a single number from a persistence diagram, while amplitudes rely on metrics made to compare diagrams. 

 \begin{table}[!t]
 \centering
 \caption{The $8$ most important uncorrelated features}
 \label{bestof_decor}
\begin{tabular}{l|l|l|l}
\textbf{Filtration} & \textbf{Parameters}          & \textbf{Dim} & \textbf{Vectorization}     \\ \hline \hline
Height     & Direction: $[1,0]$   & $0$ & Persistent entropy  \\ \hline
Radial     & Center: $(6,13)$     & $0$ & Persistent entropy  \\ \hline
Height     & Direction: $[1,-1]$  & $0$ & Persistent entropy  \\ \hline
Radial     & Center: $(0,6)$      & $0$ & Persistent entropy  \\ \hline
Density    & $\#$Neighbors: $6$   & $1$ & Persistent entropy  \\ \hline
Height     & Direction: $[1,1]$   & $0$ & Persistent entropy  \\ \hline
Height     & Direction: $[-1,0]$  & $0$ & Persistent entropy  \\ \hline
Height     & Direction: $[0,1]$   & $0$ & Persistent entropy  \\ 
\end{tabular}
\end{table}

\begin{figure}[!t]
     \centering
     \includegraphics[scale=0.45]{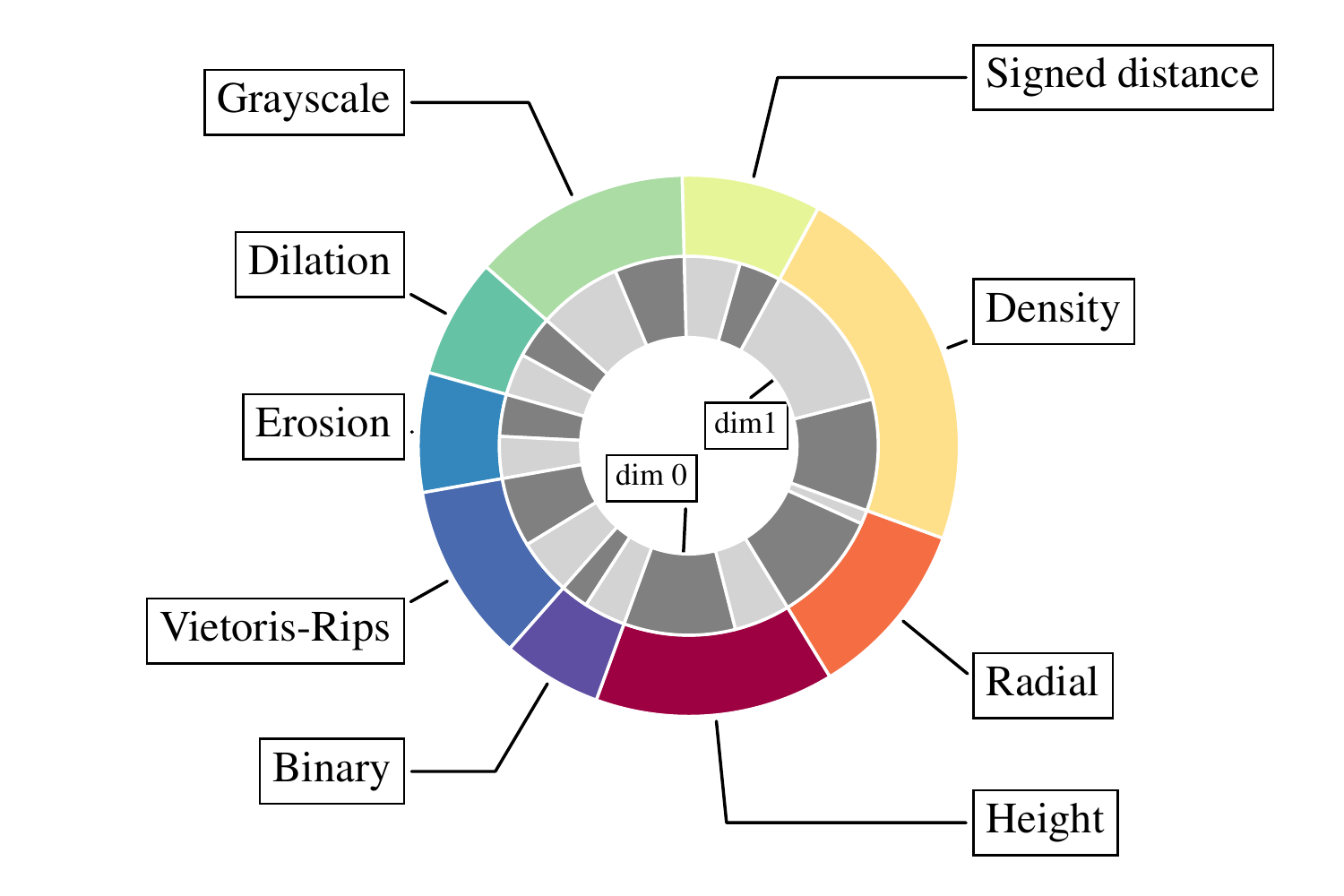}
     \caption{The percentage of each type of filtration in the $84$ uncorrelated features and the partition by dimension of the persistence diagrams.}
     \label{pie}
 \end{figure}

We highlight the filtrations and dimensions of the $84$ uncorrelated features in Figure \ref{pie}.
All filtrations are represented but the density filtration introduced in this work is proven to be especially effective. It can detect the persistence of $1-$cycles in terms of the thickness of the lines that surround them. 
Our interpretation of the grayscale filtration importance is that it highlights contours of lines first and therefore how much pressure is put on the pen while writing: it is able to differentiate when several lines are drawn in an unsmooth process. 
While not in the top $8$, the Vietoris-Rips, erosion, dilation, and signed distance filtrations still provide information on the distance between pixels and the size of holes in a binary image. In the specific case of MNIST, they might not be as useful as in other dataset where the size of holes varies more.
The binary filtration relates to homology, which intuitively can classify digits in three different classes:  $\{1,2,3,5,7\}$, $\{4,6,9\}$ and $\{8\}$, based on the number of $1$-cycles. Persistent homology, by adding information on the position and scale of those topological features, can provide more information to distinguish the digits from each other.
The percentages of dimensions $0$ and $1$ are balanced in the $84$ best features, but the  dimension $0$ features are ranked higher in term of importance. The numbers and positions of $1$-cycles are still relevant but their information can be more redundant, as subdiagrams of dimension $1$ show less diversity. 

\subsection{Supervised learning} \label{supervised}

In order to classify the digits, we apply a random forest classifier with $10,000$ trees. As a reference performance, we use the pixel values of the image as features. Several pixels are always $0$, \textit{i.e.} at the border, and therefore we only consider an effective number of $703$ pixels instead of $784$.   
\begin{figure}[!t]
    \centering
    \includegraphics[scale=1]{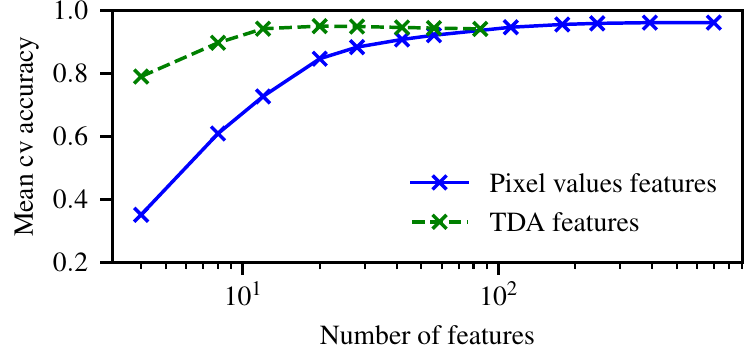}
    \caption{Cross validation means accuracy of the random forest against the number of features used for two different pipelines: the one with the pixel values as features and the one coming from decorrelated features. }
    \label{results}
\end{figure}

To show that essential information about the shape of the image is captured by a much smaller number of topological features, we train both classifiers on a changing number of features ordered by importance. The results of our experiments are shown in Figure~\ref{results}. The reference pipeline achieves a mean accuracy of $96.3 \%$~\footnote{Note that Convolutional Neural Networks typically achieve an accuracy of $100 \%$ on MNIST.} from $112$ features on. Our pipeline of selected uncorrelated topological features achieves similar accuracy using only $28$ features. Therefore, it is an effective dimensionality reduction algorithm capable of capturing highly relevant topological features of the digits.

\begin{figure}[!t]
    \centering
    \hspace{-0.2cm}\includegraphics[scale=0.40]{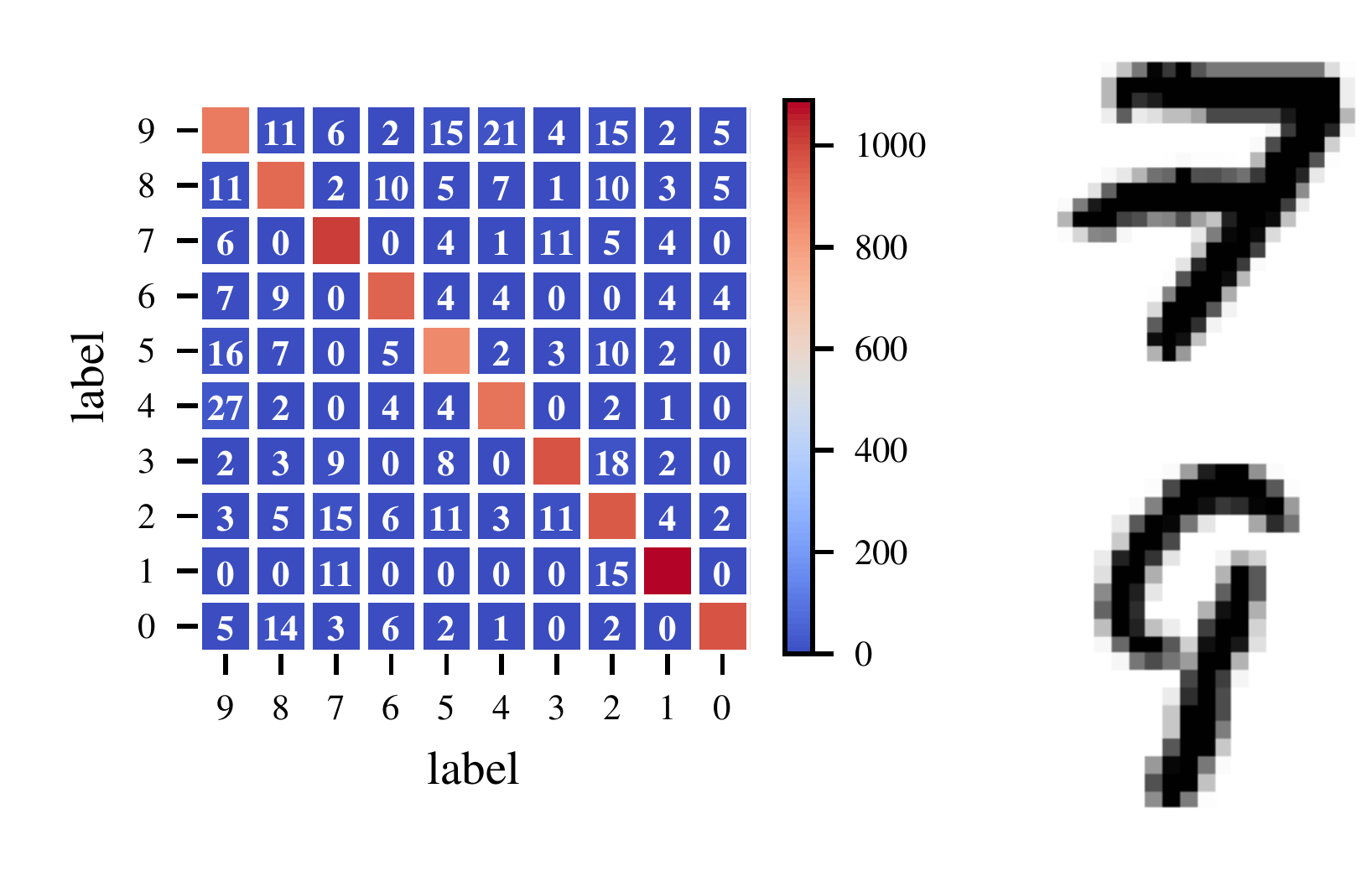}
    \put(-55,00){ \scriptsize (c) $9$ classified as a $4$.}
      \put(-55,63){ \scriptsize (b) $7$ classified as a $3$.}
      \put(-125,0){\scriptsize(a)}
    \caption{The confusion matrix $\mathcal{C}(i, \, j)$ that summarizes whenever a digits $i$ is predicted as a digit $j$, and two wrongly classified digits (b), (c). }
    
    \label{confusion}
\end{figure}

To understand where our topological pipeline of $28$ features fails, we apply it to classify the test set of the remaining $10,000$ images. We first plot the confusion matrix in Figure~\ref{confusion}a. Looking further, we investigate the common errors that our TDA pipeline does, as in Figure \ref{confusion}b, where a $7$ gets misclassified as a $3$. Even though the human eye recognizes a $7$ because of the sharp angles, the topological pipeline mostly captures the three connected components that appear on height or radial filtrations that start from the left side of the image, explaining the misclassification. The $9$ of Figure \ref{confusion}c gets misclassified as a $4$ by our pipeline because the top loop is not closed. Topological features are very sensitive to loops and disconnections and therefore it is difficult to use them alone to classify an object when it does not have its expected shape.

\section{Conclusion and further steps}

In this paper, we combined a wide range of different TDA techniques for images based on different filtrations and diagram features. We were able to classify MNIST digits using $5$ times less features than the pixel values, while maintaining the accuracy. As a result, we show that TDA techniques provide powerful dimensionality reduction algorithms for images.

Combining machine learning and TDA into a generic pipeline which gathers a wide range of TDA techniques was shown to be a powerful approach to understand the underlying characteristic shapes on the images of a dataset, especially using height, radial, and our density filtrations. Moreover, through the systematic study of feature importance, one can easily validate and support one's choice of topological features for novel datasets.

The results presented could be further improved by allowing several thresholds for the binarization of images. Indeed, these would lead to different filtrations and some $1$-cycles can be badly detected if the binarization threshold is too low or too high. Moreover, the amplitude choice can be too simplistic. Diagram metrics and kernel could be used more meaningfully, for instance, by calculating pairwise distances to the centroid of clusters in the moduli space of diagrams.

There exist many interesting further steps for this work as TDA has just begun to be used in a machine learning context. For instance, it could be useful to study and classify the same objects but at different scales. Scaling an image is directly related to scaling its persistence diagrams. An analysis of MNIST with different scales would be interesting. Similarly, with sufficiently many directions, one should be able to detect rotations quite easily, and hence classify rotated MNIST. Moreover, TDA is reputed for its robustness to noise, therefore it would be interesting to conduct the same study with different types of noise.

\section*{Acknowledgment}
The authors would like to thank Dr. Matteo Caorsi, Dr. Umberto Lupo, Dr. Anibal Medina Mardones and Dr. Gard Spreemann for all the fruitful discussions and Prof. Kathryn Hess for her careful revision of the paper. This work relies on the giotto-learn TDA library for machine learning~\cite{giotto} which makes use of the GUDHI~\cite{gudhi}, ripser~\cite{ripser} and hera~\cite{hera} TDA libraries and scikit-learn \cite{sklearn}.

\bibliographystyle{IEEEtran}
\bibliography{IEEEabrv,bibliography}
\end{document}